\documentclass[journal,peerreview,onecolumn,letterpaper]{llncs}

\usepackage{tabularx}
\usepackage{cite}
\usepackage{amsmath,amssymb,amsfonts}
\usepackage{algorithmic}
\usepackage{graphicx}
\usepackage{textcomp}
\usepackage{xcolor}
\usepackage{enumitem}
\usepackage{float}
\usepackage{booktabs}
\usepackage{subcaption}
\usepackage{caption}
\newcommand{\repeatthanks}{\textsuperscript{\thefootnote}}

\begin{document}
\title{Two-stage Synthetic Supervising and Multi-view Consistency Self-supervising based Animal 3D Reconstruction by Single Image}

\author{Zijian Kuang\inst{1}\thanks{Equal contributions} \and
Lihang Ying\inst{2}\repeatthanks \and
Shi Jin\inst{3} \and
Li Cheng\inst{4}}
\institute{Department of Computing Science, University of Alberta, Edmonton, Canada\\
\email{Email: \inst{1}kuang@ualberta.ca}\\
Zerobox Inc., Edmonton, Canada\\
\email{Email: \inst{2}leo@zerobox.ai, \inst{3}shi@zerobox.ai}\\
Department of Electrical and Computer Engineering, University of Alberta, Edmonton, Canada\\
\email{Email: \inst{4}lcheng5@ualberta.ca}}

\maketitle
\thispagestyle{plain}
\pagestyle{plain}

\begin{abstract}
Pixel-aligned Implicit Function (PIFu) effectively captures subtle variations in body shape within a low-dimensional space through extensive training with human 3D scans, its application to live animals presents formidable challenges due to the difficulty of obtaining animal cooperation for 3D scanning. To address this challenge, we propose the combination of two-stage supervised and self-supervised training to address the challenge of obtaining animal cooperation for 3D scanning. In the first stage, we leverage synthetic animal models for supervised learning. This allows the model to learn from a diverse set of virtual animal instances. In the second stage, we use 2D multi-view consistency as a self-supervised training method. This further enhances the model's ability to reconstruct accurate and realistic 3D shape and texture from largely available single-view images of real animals. The results of our study demonstrate that our approach outperforms state-of-the-art methods in both quantitative and qualitative aspects of bird 3D digitization. The source code is available at \url{https://github.com/kuangzijian/drifu-for-animals}.
\end{abstract}


\section{Introduction}
The exploration of animal detection, tracking, and behavior analysis plays a crucial role in various fields like biology, ecology, farming, and entertainment. However, in the computer vision community, the focus has predominantly been on human modeling and behavior analysis.

Acquiring 3D data for training models poses significant challenges. Models such as PIFu \cite{pifu} and SMPL \cite{SMPL:2015} heavily rely on extensive databases containing thousands of 3D scans, encompassing diverse human shapes and poses. Humans, being cooperative subjects, make this process more feasible. Unfortunately, capturing multiple wild animals for controlled scanning in a lab setting is impractical, and the logistics involved in taking scanning equipment into the wilderness are complex.

To tackle this challenge, we propose a two-stage model that combines Differentiable Rendering and Implicit representation. This innovative approach leverages synthetic animal 3D models to train the model and enhances its capabilities by generating 3D models from single-view images of real animals.

In Stage 1, our process utilizes a pixel-aligned implicit function to predict the continuous inside/outside probability field of a synthetic bird based on the provided image. Using advanced differentiable rendering techniques, we create a render for the 3D implicit representation of the synthetic bird generated by a pixel-aligned feature decoder. This rendered output is then transformed into 2D images, facilitating multi-view self-supervised learning.

Moving to Stage 2, the model incorporates a pixel-aligned feature encoder-decoder, pre-trained on synthetic birds. To enhance the model's adaptability to real-world scenarios, we employ transfer learning by integrating real bird images along with their silhouettes. This two-stage strategy ensures a robust and versatile approach to animal 3D model generation, where synthetic data aids in the learning process and real-world images contribute to the model's practical utility and generalization capability.

Our experiment primarily focuses on the 3D digitization of birds. For Stage 1, we collected 20 synthetic 3D bird models representing various bird types, such as owls, blue jays, toucans, parrots, ducks, and pigeons, to train our model. In Stage 2, we used 5964 previously unseen real bird images, along with their silhouettes, for additional model training.

The results of our study demonstrate that our differentiable rendering and implicit function-based approach outperforms state-of-the-art methods \cite{19,UMR} in both quantitative and qualitative aspects of bird 3D digitization. Furthermore, we extended our method to other animals, including horses, cows, bears, and dogs, with qualitative results showcased in this paper.

The main contribution of this work is the combination of two-stage supervised and self-supervised training to address the challenge of obtaining animal cooperation for 3D scanning. In the first stage, we leverage synthetic animal models for supervised learning. This allows the model to learn from a diverse set of virtual animal instances. In the second stage, we use 2D multi-view consistency as a self-supervised training method. This further enhances the model's ability to reconstruct accurate and realistic 3D shape and texture from largely available single-view images of real animals.

\section{Related Work}
\label{RW}
\subsection{3D Shape Representation}
Researchers have explored various representations for 3D processing tasks, including point clouds \cite{6}, implicit surfaces \cite{25,26}, triangular meshes \cite{19, 21, 24, 20, 38, 27, 39}, and voxel grids \cite{3, 8, 11, 36, 40, 45, 50, 12}. While both voxels and triangular meshes are suitable for deep learning architectures (e.g., VON \cite{41, 49}, PointNet \cite{29, 30}), they face challenges such as memory inefficiency or limited differentiable rendering capabilities. Therefore, in this study, we adopt point clouds \cite{19, 21, 24, 20, 38, 27, 39} as the preferred 3D shape representation for reconstruction tasks.

\subsection{Single-view 3D Reconstruction}
The objective of single-view 3D reconstruction \cite{pifu,3, 8, 11, kuang2022normalizing, 36, 40, 45, 50, 6, 14} is to generate a three-dimensional shape from a single input image. This challenging task has been approached by various methods with different levels of supervision. Some approaches \cite{38, 27, 39, sun2022product} rely on paired image and ground truth 3D mesh data, which requires extensive manual annotation efforts \cite{43} or is limited to synthetic data \cite{1}. More recent approaches \cite{21, 24, 20, 2, kuang2022funet} mitigate the need for 3D supervision by leveraging differentiable renderers \cite{21, 24, 2} and adopting the "analysis-by-synthesis" approach, either using multiple views or known ground truth camera poses. In order to alleviate supervision constraints, Kanazawa et al. \cite{19} explored 3D reconstruction using a collection of images depicting different instances. However, their method still relies on annotated 2D keypoints to accurately infer camera pose. This work is also significant as it introduces the concept of a learnable category-level 3D template shape, although it requires initialization from a keypoint-dependent 3D convex hull. Similar problem settings have been investigated in other methods \cite{33, 42, 15}, but they are limited to rigid or structured objects like cars or faces. In contrast, our approach encompasses both rigid and non-rigid objects (e.g., birds, horses, penguins, motorbikes, and cars). We propose a method that jointly estimates the 3D shape and texture from a single-view image, utilizing a synthetic animals' 3D template and a collection of real animal images with silhouettes as supervision. Essentially, we eliminate the need for real animals' 3D template priors, annotated key points, or multi-view real animal images.

\section{Approach}
\label{PM}

\begin{figure}[H]%
\centering
\includegraphics[width=1\textwidth]{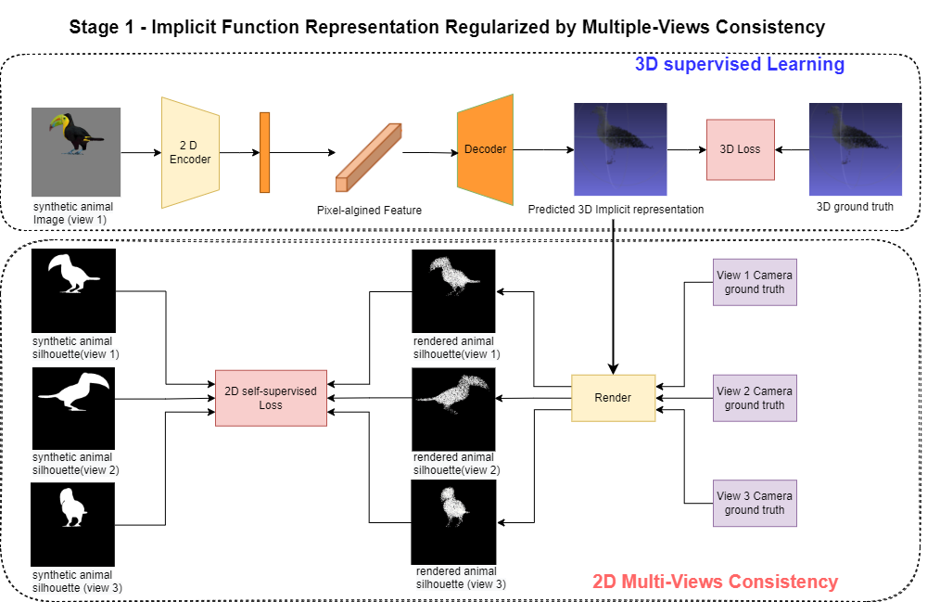}
\caption{Overview of our pipeline for digitalizing models using differentiable rendering and implicit functions. In Stage 1, we use an implicit function to predict the continuous probability field \cite{pifu} for the synthetic bird's inside/outside regions based on the given bird image. Then, using differentiable rendering, we generate a render of the 3D implicit representation of the synthetic bird produced by a pixel-aligned feature decoder, rendering it into 2D images for multi-view self-supervised learning.}
\label{fig1}
\end{figure}

In order to reconstruct the 3D mesh of an object instance from an image, a network needs to be able to predict the shape, texture, and camera pose simultaneously. To accomplish this, we use the existing network introduced in \cite{pifu} (PIFu) as the foundation for our reconstruction network. Our goal is to accurately reconstruct the animal's 3D geometry and texture from single or multi-view images, while preserving the details shown in Figure \ref{fig1}.

\subsection{2D multi-view consistency}

We aim for the 3D implicit representation to exhibit uniformity from different viewing angles. To achieve this, we employ a 2D multi-view consistency strategy, utilizing differentiable rendering techniques as depicted in Figure \ref{fig1}. This approach ensures that the pixel-level implicit function gains additional insights from synthetic animal 3D models. By applying a render function \(R(P)\), where \(P\) represents the predicted point cloud 3D implicit representation, the model generates three different views \(I_v\) by adjusting camera parameters. This enables 2D self-supervised learning through MSE loss specifically on the rendered 2D views.
\begin{equation}
\mathcal{L}_M=\frac{1}{n} \sum_{i=1}^n\left(I_{vi}-\hat{I}_{gi}\right)^2
\end{equation}

where \(n\) represents the number of sampled points. During training, we choose to use a point cloud as the 3D implicit representation instead of marching cubes. The reason for this choice is that the marching cubes algorithm is not differentiable, which makes it difficult to optimize with gradients during inference.

The traditional Marching Cubes algorithm is a non-differentiable geometric method used for surface reconstruction and visualization of isosurfaces from 3D scalar fields. It involves thresholding the scalar field and constructing polygonal surfaces based on the intersections of a binary mask with a grid of cubes.

To enable differentiable operations and facilitate gradient-based optimization, differentiable versions of Marching Cubes or similar algorithms are used in applications such as differentiable rendering and neural network training.

In our approach, we use differentiable rendering methods to create a render of the 3D implicit representation of the synthetic bird generated by a pixel-aligned feature decoder. We render this point cloud 3D representation into three fixed camera views (0 degrees, 90 degrees, and 180 degrees) for self-supervised learning.

\begin{figure}[H]%
\centering
\includegraphics[width=1\textwidth]{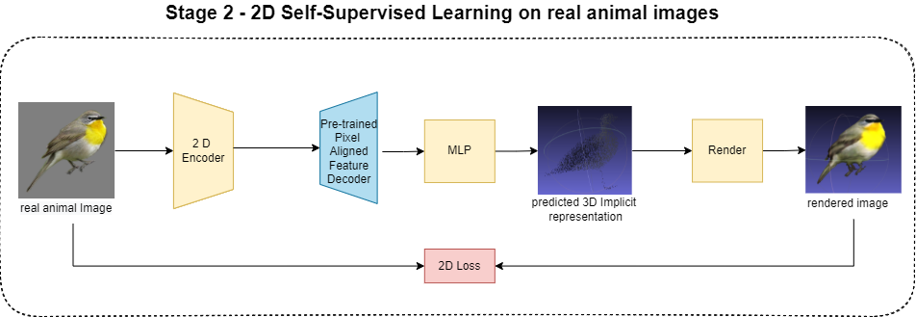}
\caption{In Stage 2, we utilize a pre-trained pixel-aligned feature encoder-decoder that was trained on synthetic birds. We incorporate real bird images and their silhouettes through transfer learning.}
\label{fig2}
\end{figure}

\subsection{2D self-supervised learning on real animal images}
In the second stage of our approach, we fine-tune the pre-trained implicit function on synthetic animal images by incorporating real animal images using a self-supervised learning approach, as depicted in Figure \ref{fig2}. Initially, the implicit function generates a 3D representation, which is then rendered into a 2D image using the render function \(R(P)\). We apply a 2D loss function to compare the rendered 2D image with the ground truth image. During inference, the trained implicit function can be directly applied to a given 2D animal image within the trained classes from both stages, specifically for birds in this case.

\section{Experiments}
We conducted experiments to evaluate the effectiveness of our proposed methodology on diverse datasets. These datasets included the CUB-200-2011 dataset \cite{cub}, which consists of bird images, and the ImageNet dataset \cite{imagenet}, which includes images of horses, zebras, and cows, encompassing a wide range of animal species.

\subsection{Implementation Details}\label{subsec1}

In our implementation, we utilized the PyTorch3D library \cite{ravi2020accelerating}, which provides differentiable rendering capabilities for rasterizing batches of point clouds.

For Stage 1, we trained the 3D implicit function network and the 2D render network for 100 epochs with a learning rate of 0.001. In Stage 2, we further fine-tuned the refined implicit function network for 50 epochs with a learning rate of 0.0005.

\begin{figure}[H]%
\centering
\includegraphics[width=1\textwidth]{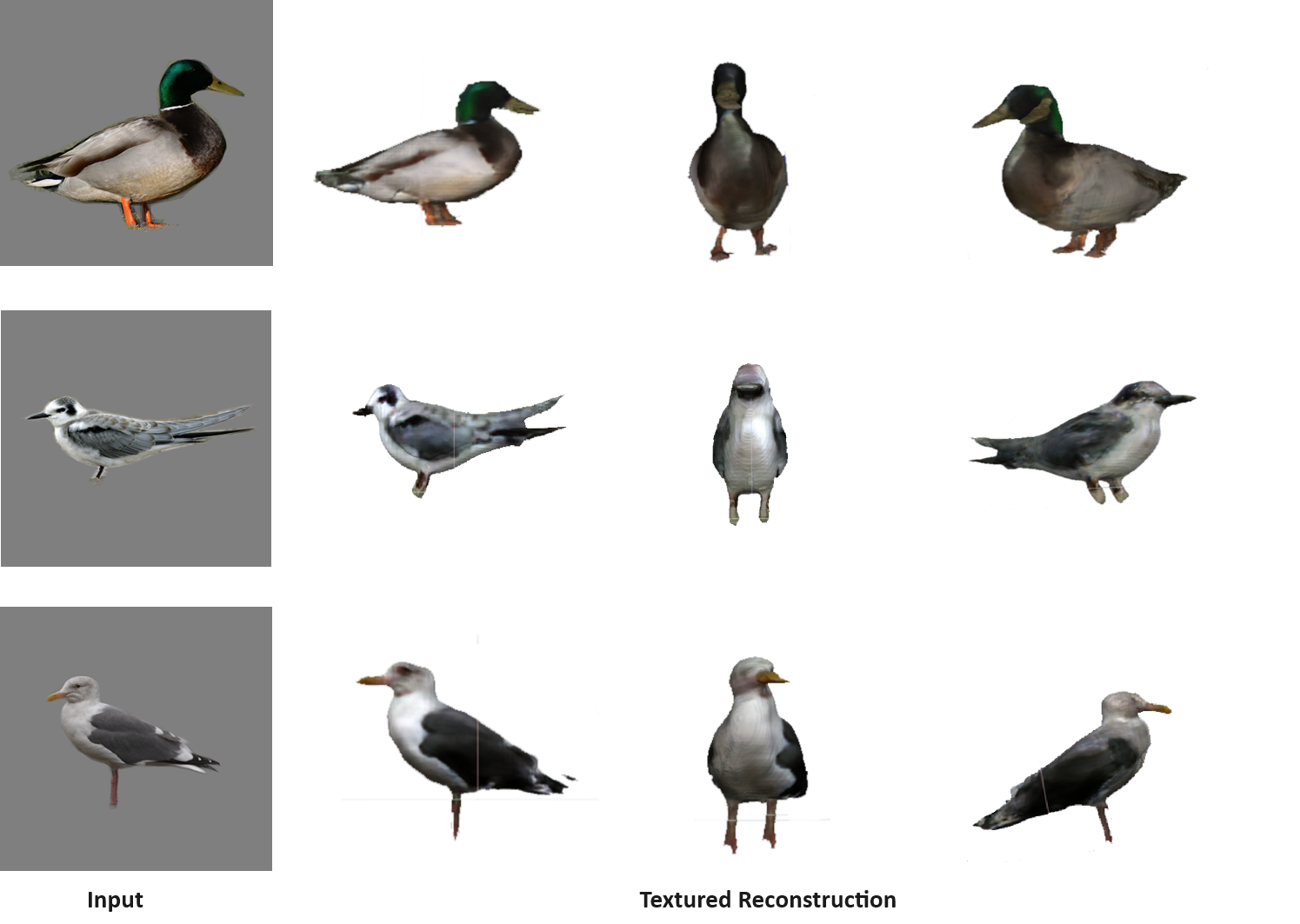}
\caption{Qualitative results showcasing single-view 3D and textured reconstructions of real bird images from the CUB-200-2011 dataset \cite{cub}.}
\label{fig3}
\end{figure}

\subsection{Qualitative Results}\label{subsec5}
\begin{figure}[H]%
\centering
\includegraphics[width=0.8\textwidth]{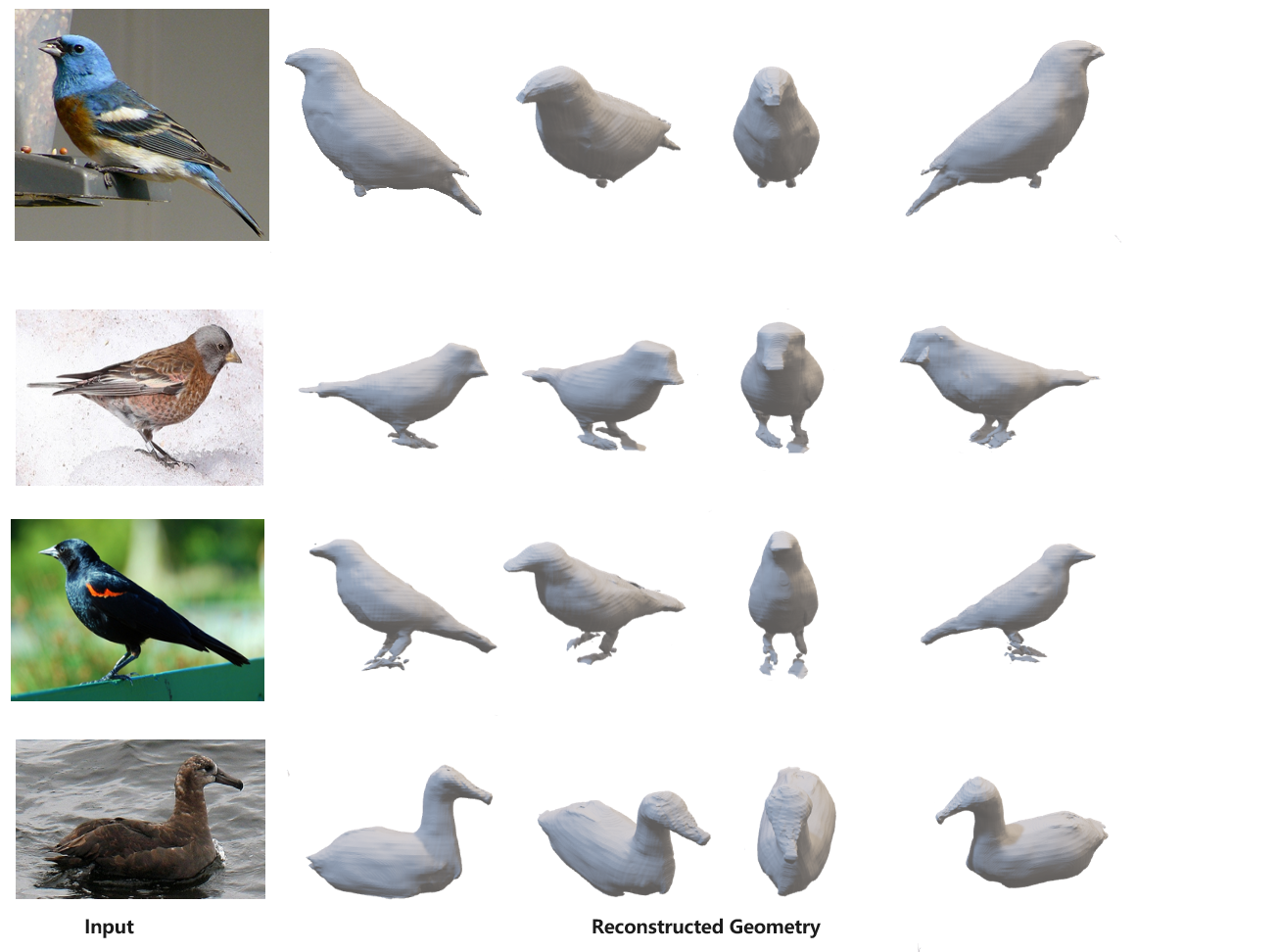}
\caption{Qualitative single-view 3D reconstruction results on real bird images from the CUB-200-2011 dataset \cite{cub} are shown in Figure \ref{fig4}.}\label{fig4}
\end{figure}

\begin{figure}[H]%
\centering
\includegraphics[width=0.8\textwidth]{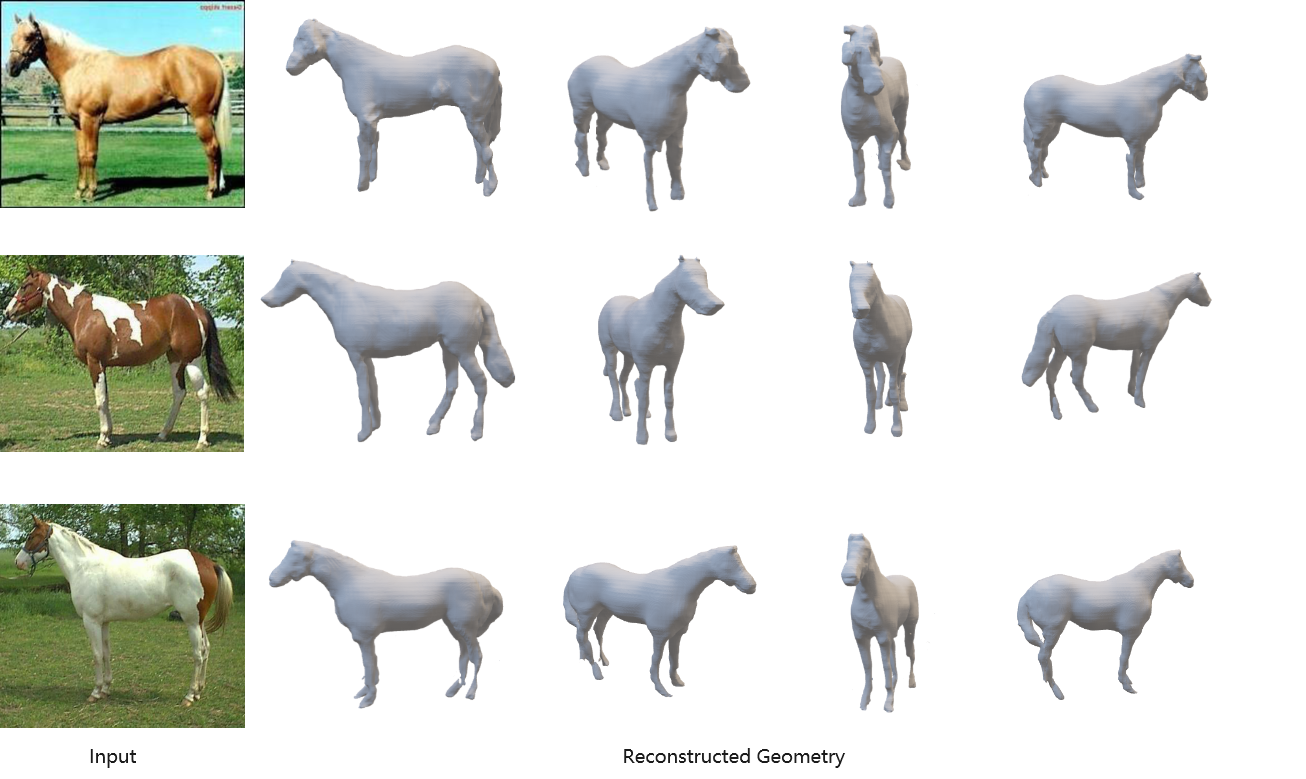}
\caption{Qualitative single-view 3D reconstruction results on real animal images from the Weizmann horses dataset \cite{weizmann} can be seen in Figure \ref{fig5}.}\label{fig5}
\end{figure}

To qualitatively evaluate our model, we present the 3D representations of birds and horses in Figure \ref{fig3}, Figure \ref{fig4}, and Figure \ref{fig5}. These figures highlight the distinctive shape features of each category. Remarkably, our model excels in horse prediction, outperforming CMR and UMR by accurately capturing intricate details such as the legs and tail. The visualizations vividly demonstrate the remarkable accuracy and level of detail achieved by our model.

Figure \ref{fig3} showcases the results of our digitization process using real bird images from the CUB-200-2011 dataset \cite{cub}. Our DRIFu model demonstrates its adaptability across a diverse range of bird species, generating high-resolution local details and inferring plausible 3D surfaces, even in previously unseen regions. Moreover, it successfully infers complete textures from a single input image, providing a comprehensive view of the 3D models from various angles.

For evaluating texture reconstruction, we employed precision and recall metrics, comparing the rendered 2D images with ground truth images. Our approach outperforms other models in terms of texture assessment as well.

\subsection{Quantitative Results}\label{subsec6}
We evaluate our reconstruction accuracy quantitatively using three metrics for 2D shape and texture. However, the lack of ground truth for 3D meshes or camera poses in these datasets makes a quantitative evaluation in 3D challenging.

We first assess the shape reconstruction for birds. Since the CUB-200-2011 dataset \cite{cub} does not provide ground truth 3D shapes, we follow the approach of \cite{19} and calculate the mask reprojection accuracy using the intersection over union (IoU) between the rendered and ground truth silhouettes. Table.~\ref{tab1} demonstrates that our model outperforms other state-of-the-art single-view bird 3D reconstruction models, indicating its ability to predict 3D mesh reconstructions that align well with 2D observations. 

Similar to the shape evaluation, texture assessment is performed by comparing the rendered and ground truth 2D images using precision and recall measures. Our approach also surpasses the other models in terms of texture accuracy.

\begin{table}[h]
\caption{Quantitative evaluation of mask IoU and texture precision and recall on the CUB-200-2011 dataset \cite{cub}. The comparisons are made against the baseline supervised models \cite{19,UMR}.}\label{tab1}%
\begin{tabular}{@{}llll@{}}
\toprule
Methods & Mask IoU  & Texture Precision & Texture Recall\\
\midrule
CMR \cite{19}    & 0.706   & 0.652  & 0.612  \\
UMR \cite{UMR}    & 0.734   & 0.701  & 0.54  \\
Ours    & \textbf{0.836}   & \textbf{0.837}  & \textbf{0.721}  \\
\end{tabular}
\end{table}

We also conducted additional evaluations comparing our model with the baseline supervised models \cite{19,UMR} using real horse images from the Weizmann Horses dataset \cite{weizmann}. The results of this evaluation are presented in Table.~\ref{tab2}. The use of implicit representation enables us to achieve higher fidelity in the reconstruction.

\begin{table}[h]
\caption{Quantitative evaluation of mask IoU and texture precision and recall on the Weizmann Horses dataset \cite{weizmann}. The comparisons are made against the baseline supervised models \cite{19,UMR}.}\label{tab2}%
\begin{tabular}{@{}llll@{}}
\toprule
Methods & Mask IoU  & Texture Precision & Texture Recall\\
\midrule
CMR \cite{19}    & 0.606   & 0.552  & 0.423  \\
UMR \cite{UMR}    & 0.534   & 0.601  & 0.54  \\
Ours    & \textbf{0.765}   & \textbf{0.801}  & \textbf{0.721}  \\
\end{tabular}
\end{table}
\section{Conclusion}
The goal of this research is to develop a model that can reconstruct the 3D shape and texture of animals using single-view images of real animals. One key aspect of our approach is the use of synthetic animal 3D models for supervision, allowing the model to learn from a diverse set of virtual animal instances. A crucial part of our methodology is the introduction of a self-supervised differentiable rendering framework, which ensures consistency between the reconstructed 3D representations and the corresponding images. This approach reduces ambiguities in predicting 3D shape and texture from 2D observations.

In addition, we incorporate a transfer learning-based self-supervised framework, which enables our model to leverage the learned 3D representation when dealing with unseen real animal images. This transfer learning mechanism enhances the adaptability and robustness of our model in real-world scenarios.

The effectiveness of our proposed method is demonstrated through extensive experiments. Notably, our approach outperforms state-of-the-art supervised category-specific reconstruction techniques. These results highlight the potential and versatility of our model in advancing the field of 3D shape and texture reconstruction from single-view images, particularly for real animals.

\bibliographystyle{IEEEtran}
\bibliography{main}

\end{document}